\documentclass[11pt,a4paper]{article}
\usepackage[hyperref]{naaclhlt2019}
\usepackage{times}
\usepackage{graphicx}
\usepackage{latexsym}
\usepackage{booktabs}
\usepackage{csquotes}
\usepackage{comment}
\usepackage{amssymb}
\usepackage{amsmath}
\usepackage{fancyvrb}
\usepackage{xcolor}
\usepackage{subcaption}
\usepackage{adjustbox}
\usepackage{array}
\usepackage{rotating}
\usepackage{multirow}
\newcolumntype{R}[2]{%
    >{\adjustbox{angle=#1,lap=\width-(#2)}\bgroup}%
    l%
    <{\egroup}%
}
\newcommand*\rot{\multicolumn{1}{R{45}{1em}}}

\definecolor{cadmiumgreen}{rgb}{0.0, 0.42, 0.24}

\usepackage{url}
\usepackage{verbatimbox}

\aclfinalcopy % Uncomment this line for the final submission
\newcommand{\comk}[1]{{\color{orange} #1}}

\title{Automatic Accuracy Prediction for AMR Parsing}

\author{Juri Opitz {\normalfont and} Anette Frank \\
  Research Training Group AIPHES\\
  Leibniz ScienceCampus ``Empirical Linguistics and Computational Language Modeling''\\
  Department for Computational Linguistics \\
  69120 Heidelberg \\
 {\tt $\lbrace$opitz,frank$\rbrace$@cl.uni-heidelberg.de} }

\date{}

\begin{document}
\maketitle
\begin{abstract}
Abstract Meaning Representation (AMR) re\-presents 
%each 
sentences as 
%a 
directed, acyclic and rooted graphs,
%and aims
aiming at capturing 
%a sentence's 
their meaning in a machine readable format. AMR \textit{parsing} converts 
%a 
natural language sentences into such graphs. However, evaluating a parser on new data by means of comparison to manually created AMR graphs is very costly. Also, we would like to be able to detect parses of questionable quality, or preferring results of alternative systems by selecting the ones for which we can assess good quality.
We propose %and define 
AMR \textit{accuracy prediction} as the task of predicting several metrics of correctness for an automatically generated AMR parse -- in absence of the corresponding gold parse. %For solving this task 
We develop a neural end-to-end multi-output regression model and perform three %different 
case studies: firstly, we evaluate 
%its capacity to automatically predict a large suite of AMR accuracy metrics.
the model's capacity of predicting AMR parse accuracies and test whether it can reliably assign high scores to gold parses. % with respect to the gold scores.
Secondly, we perform parse selection based on %individual ranked 
predicted parse accuracies of candidate parses from alternative systems, with the aim of 
%improved 
improving overall results. 
%And 
Finally, we predict system ranks for submissions from two AMR shared tasks on the basis of their 
%averaged 
predicted parse accuracy averages. All experiments are carried out across two different domains and show that our method is effective. 
\end{abstract}

\section{Introduction}
Abstract Meaning Representation (AMR) \cite{Banarescu13abstractmeaning} represents the semantic structure of a sentence, including concepts, semantic operators and relations, sense-disambiguated predicates and their arguments. % (semantic roles).  
\begin{figure}
%\begin{verbatim}
\begin{verbnobox}[\small]
(a / asbestos               
  :polarity -               
  :time (n / now)           
  :location (t / thing      
    :ARG1-of (p / produce-01  
    :ARG0 (w / we))))
\end{verbnobox}
%\end{verbatim}
\caption{
Humanly produced AMR for: \textit{There is no asbestos in our products now.}
%The numbers of 
Numbered predicates refer to PropBank senses \cite{Palmer:2005:PBA:1122624.1122628}.}
\label{fig:ex1}
\end{figure}
\begin{comment}
\begin{figure}
\begin{verbnobox}[\small]
(a / asbestos             
  :time (n / now)
  :polarity -
  :location (p / product
    :poss (w / we)))

Smatch: 0.70, Negation: 1.00, SRL: 0.00, COncepts: 0.667
        
(a / asbestos
  :ARG1 (w / we
  :ARG1-of (p / product
    :mod (n / now)))
    :polarity -)
    
Smatch: 0.30, Negation: 0.00, SRL: 0.14, 0.444

(a / asbestos
    :polarity -
    :location (p / product)
    :time (n / now))
    
Smatch: 0.667, Negation; 1.00, SRL 0.00, Concepts: 0.5
\end{verbnobox}
\caption{\comk{Three different AMR parses for the sentence \textit{There is no asbestos in our products now.}}
GPLA (middle), JAMR (bottom).}
\label{fig:ex2}
\end{figure}
\end{comment}
As
%Since AMR provides 
a machine readable representation of the meaning of a sentence, 
AMR
%it 
is potentially useful for many 
%different 
NLP tasks. Among other applications
%cases 
it has been used in machine translation \cite{DBLP:conf/coling/JonesABHK12}, text summarization \cite{DBLP:journals/corr/abs-1805-10399,DBLP:journals/corr/DohareK17} and question answering \cite{AAAI1612345}. Since the introduction of AMR, 
%a considerable number of
many approaches to AMR parsing have been proposed: graph-based pipeline systems which rely on an alignment step \cite{P14-1134,DBLP:conf/semeval/FlaniganDSC16} or 
%and 
transition-based parsers relying on dependency annotation \cite{wang-xue-pradhan:2015:NAACL-HLT,wang-xue-pradhan:2015:ACL-IJCNLP,wang-EtAl:2016:SemEval}. In the following we will denote the former by \textbf{JAMR} and the latter by \textbf{CAMR}. More recently, end-to-end neural systems have been proposed which produce linearized AMR graphs %either in a 
within character-based \cite{DBLP:journals/corr/NoordB17a} or word-based \cite{DBLP:journals/corr/KonstasIYCZ17} encoding models. %Due to scarce training data, 
Both approaches 
greatly profit from
large 
amounts of
%additional 
silver training data.  
The silver data is obtained
%to boost their performances. This is done either 
%via a specially designed 
with self-training 
%cycle 
\cite{DBLP:journals/corr/KonstasIYCZ17} or 
%by 
the aid of additional parsers, where only parses with considerable agreement
%an agreement above a certain threshold 
are chosen to extend the training data \cite{DBLP:journals/corr/NoordB17a}. \citet{DBLP:journals/corr/abs-1805-05286} %approach the task as a graph prediction problem and 
formulate a neural model that
%for 
jointly predicts
%ing 
alignments, concepts and relations. Their system -- henceforth 
%denoted as
called %achieves 
\textbf{GPLA} (Graph Prediction with Latent Alignments) -- defines
the current state-of-the-art in AMR parsing. 
%result 
%on \comk{the} two main data sets 

A system that can perform accuracy prediction for AMR parsing can be used in a variety of ways: (i) \textit{estimating the quality of downstream tasks} %\comk{that deploy AMR parses}. 
that deploy AMR parses.
E.g., in a document summarization scenario, we might expect lower %summary 
quality %\comk{of a summary} 
of a summary if the estimated quality of AMR parses used %\comk{as a basis for the summary} 
as a basis for the summary
%by the summarization system 
is low; (ii) 
%applying 
AMR parsing 
accuracy
%quality 
estimation %\comk{can be used to produce \textit{high-quality automatically parsed data}:} 
can be used to produce \textit{high-quality automatically parsed data}:
%\comk{by fitering} 
by filtering
%to 
the outputs of single parsing systems in self-training, 
%or 
%\comk{by selecting} 
by selecting
%to select 
high-quality outputs from
different parsing systems 
in a tri-parsing setting, or else by predicting overall rankings over alternative parsing systems applied to in- or out-of-domain data; (iii) finally, AMR parse accuracy prediction could be used in the context of a \textit{parser-supported treebank
construction} process. 
E.g.,
%For example,
in an active learning scenario, we can select useful targets for manual annotation based on their expected efficiency for parser improvement -- the fine-grained evaluation measures predicted by our system can be used for targeted improvements. 
%Or, i
In the simplest case, we can provide the human annotator with automatic parses where only few flaws have to be mended. Hence, AMR accuracy prediction systems have the potential to tremendously reduce manual annotation cost and time.

\paragraph{Contributions} %\textcolor{blue}{maybe reformulate: coarser} 
We define AMR accuracy prediction as the task of predicting a rich suite of metrics to assess various subtasks covered by AMR parsing (e.g.\ negation detection or semantic role labeling). To approach this task, we use the AMR evaluation suite suggested by \citet{DBLP:journals/corr/DamonteCS16} and develop a hierarchical multi-output regression model for %predicting \comk{12} 
%overall 36 
%different metrics 
automatically performing evaluation of 12 different tasks involved in AMR parsing (Sections \S \ref{sec:accpred} and \S \ref{sec:model}; our code is publicly accessible\footnote{\url{https://gitlab.cl.uni-heidelberg.de/opitz/quamr}}). We perform experiments in three different scenarios on unseen in-domain and out-of-domain data and show that our model (i) is able to predict scores with significant correlation to gold scores and (ii) can be used to rank \textit{parses} on a \textit{sentence-level} or to rank \textit{parsers} on a \textit{corpus-level} (\S \ref{sec:experiments}). 

\section{Related Work}
\label{sec:relwork}

Automatic accuracy prediction for %syntax
syntactic parsing comes closest to what we are doing. %The goal is to predict %\comk{the parse}
%the parse accuracy metrics given  only a sentence and its
%a
%parse tree, without any reference. 
%For example, 
\citet{ravi2008automatic}  propose a feature-based SVM regression model with RBF kernel that predicts syntactic parser performance %on data from 
on different domains. Like us, they aim at a cheap  
and effective means for estimating a parser's performance. However, in contrast to 
%this previous 
their work, our method is \textit{domain} 
and
%as well as 
\textit{parser} agnostic: we do not take into account characteristics of the domains of interest and %we also 
do not provide any performance statistics of the competing parsing systems as features to our regressor. 
\citet{PredictingthePerformanceofParsingwithReferentialTranslationMachines} addresses the task without any domain-dependent features, which results in a lower correlation to gold scores -- even if additional features from a background language model are incorporated. In contrast to the prior systems that predict a single score, we predict an ensemble of metrics suitable for assessing AMR parse quality with respect to different linguistic aspects. Also, our system does not rely on externally derived features or complex pre-processing. % and we do not predict one score but an ensemble of scores, suitable for assessing an AMR graph with respect to various subtasks involved in AMR parsing. Moreover, %as %already 
%indicated above, 
Moreover, an AMR graph differs in important ways from a syntactic tree. %structure  
Nodes in AMR do not explicitly correspond
%of the graph 
to words (as in dependency trees) or phrases (as in constituency trees). 
%The nodes 
AMR structure elements can
%be able to 
exist without any alignment to words in the sentence. To our knowledge, we are the first to propose an accuracy prediction model for AMR parsing, and offer the first general end-to-end parse accuracy prediction model that predicts an ensemble of scores for different linguistic aspects.%(to quote the designers: \enquote{someone who creates an AMR from English may not provide links between AMR concepts and English word tokens}\footnote{https://github.com/amrisi/amr-guidelines/blob/master/amr.md}). %We provide the system shallow pointers from nodes to tokens.

Automatic accuracy prediction 
%for NLP systems 
has also been researched for PoS-tagging \cite{VanAsch:2010:UDS:1870526.1870531} and in machine translation. For example, \citet{soricut2012combining} predict BLEU scores for machine-produced translations. Under the umbrella of \textit{quality estimation} researchers try to predict, i.a., the post-editing time or missing words 
%of
in an automatic translation \cite{DBLP:conf/acl/CaiK13,Joshi:2016:QEE:2905055.2905259,DBLP:conf/amta/ChatterjeeNTBS18,Kim:2017:PNQ:3141228.3109480,Specia13quest-}. 
The fact that manually creating an AMR graph is significantly more costly than a translation
%graph 
%(creating an AMR graph requires %needs 
%trained linguists and takes on average 8 to 13 minutes per graph
 provides another %appealing 
compelling
argument for 
investigating
%why 
automatic AMR accuracy prediction techniques 
%are needed 
.\footnote{Creating an AMR graph requires %needs 
trained linguists and takes on average 8 to 13 minutes, cf.\ \citet{Banarescu13abstractmeaning}}

In recent work,
\citet{W11-2929,DBLP:conf/tlt/DickinsonS17,10.1007/978-3-319-18111-0_17,rehbein:ruppenhofer:2018} detect annotation errors in automatically 
%annotated
produced dependency parses.
%trees. 
The latter approach uses 
%an approach based  on  
active learning and ensemble parsing in  combination  with variational inference. 
They predict edge labelling and attachment errors and use a back-and-forth encoding mechanism from non-structured to structured tree data in order to provide the variational inference model with the needed unstructured data. 
Their work differs from ours in
three important aspects: firstly, they
%explicitly 
predict errors in 
%an
specific edges or nodes, while we %level, we 
predict an accuracy score over
%on a 
the complete graph. Moreover, our model does not \textit{need} several candidate parses as input % -- from one parse alone it can predict its accuracy 
-- when several multiple parses are available, our model can be exploited for ranking (cf.\ Sections \S \ref{subsec:app1} \& \S \ref{subsec:app2}). Finally, our method is independent of live human feedback.

\section{Accuracy Metrics for AMR Parsing}
\label{sec:accpred}
\begin{figure}
\begin{scriptsize}
\begin{Verbatim}[commandchars=\\\{\}]
(a / asbestos            (a / asbestos     
  :time (n / now)           :polarity -
  :polarity -               :location (p / product)    
  :location (p / product    :time (n / now))
    \textcolor{orange}{:poss} (w / we)))
                           __________________________
(a / \textcolor{red}{asbesto}                metr.(F1)| GP   JA   CA |             
  :polarity -               ---------|--------------|
  \textcolor{red}{:ARG1 (w / we}             Smatch   | 70 | 30 | 67 |          
    \textcolor{red}{:ARG1-of (p / product}   SRL      |  0 | 14 |  0 |
      \textcolor{red}{:mod (n / now)))})     Concepts | 67 | 44 | 50 |
                            IgnVars  | 55 |  0 | 60 |
\end{Verbatim}
\end{scriptsize}
\caption{
Three AMR parses 
%with errors 
for: 
%the sentence 
\textit{There is no asbestos in our products now}, 
%as 
generated
%with three different automatic meaning representation parses and example errors: 
%GPLA 
%\textbf{GPLA}}
%(top, \citet{DBLP:journals/corr/abs-1805-05286}), JAMR (middle, \citet{DBLP:conf/semeval/FlaniganDSC16}), CAMR (bottom, \citet{wang-xue-pradhan:2015:NAACL-HLT,wang-xue-pradhan:2015:ACL-IJCNLP,wang-EtAl:2016:SemEval}). Light errors (\textcolor{orange}{orange}) 
by GPLA
(top), JAMR (bottom),
%and 
CAMR (right). %\citet{wang-xue-pradhan:2015:NAACL-HLT,wang-xue-pradhan:2015:ACL-IJCNLP,wang-EtAl:2016:SemEval}). 
\textcolor{orange}{Light}
%are contained in the first parse, 
and \textcolor{red}{severe} errors %(\textcolor{orange}{orange}, \textcolor{red}{red}) 
are
%are %contained 
found in GPLA and JAMR 
%second 
parses;
%, while 
%the 
CAMR
%is incomplete: it
%and 
fails to provide \textit{we}, the 
%entity which is 
manufacturer of the product. Bottom right: F1 for Smatch and three example subtasks from evaluation against the gold parse (given in Figure \ref{fig:ex1}).
}
\label{fig:ex2}
\end{figure}
%\end{comment}

Automatic AMR parses are often deficient. Consider the examples in Figure \ref{fig:ex2}.  All parsers correctly detect
%ed 
the 
negation and its scope. The 
%GPLA 
GPLA parse (top) 
%does many things right and 
provides a graph structure close to the 
%human 
gold annotation (Figure \ref{fig:ex1}). However,
%GPLA 
it 
%\comk{does not correctly analyze the possessive \textit{our (product)}, which in the gold parse is represented as an object produced by the speaker (\textit{we}). It instead recognizes a location in the speaker's possession.}
does not correctly analyze the possessive \textit{our (product)}, which in the gold parse is represented as an object produced by the speaker (\textit{we}). Instead it recognizes a location in the speaker's possession. JAMR 
%parse 
(middle) fails to detect
%make out 
the concept in focus (\textit{asbestos}), possibly due to a false-positive stemming mistake. Moreover, %ignoring the stemming error,
it fails to represent that 
%the 
\textit{asbestos} is (not) in the product:
%(
it misses the \textit{:location}-edge from \textit{asbestos} to \textit{product}. 

\paragraph{AMR accuracy metrics}  
Usually, a predicted AMR graph $G$ is evaluated against a gold graph $G'$ using triple matching based on a maximally scoring variable mapping. For 
finding
%figuring out 
the optimal variable mapping, Integer Linear Programming (ILP) can be used in the Smatch metric \cite{DBLP:conf/acl/CaiK13}, which produces precision, recall and F1 score between $G$ and $G'$. While it is important to obtain a global measure of parse accuracy, we may also be interested in a quality assessment that focuses on specific subtasks or meaning aspects, 
%score, 
such as
%, for example, the 
entity linking,
%performance, 
negation detection or
%and 
word sense disambiguation (WSD). For example, if a parser commits a WSD error %(e.g., assigning the sense dive.02 instead of dive.01), 
this might be less harmful than e.g., failing to capture negation, or missing or wrongly predicting a semantic role. However, the Smatch calculation would treat 
many of such errors with equal weight --
%the different predicate and \textit{dive-02} however both 
%as equally distinct concepts from \textit{dive-01}, 
a property which in some cases may be undesirable. 

To alleviate this issue, \citet{DBLP:journals/corr/DamonteCS16} proposed an extended AMR evaluation suite which allows parser performance inspection with regard to 11 additional subtasks captured by AMR.\label{sec:tasks} %This enables us to evaluate a parser with respect to 
%distinct representational aspects of the AMR graph
%the individual subtasks of AMR 
%more closely. 
In total, 36 metrics can be computed (precision, recall and F1 for 12 tasks). F1 scores for three example metrics are displayed in Figure \ref{fig:ex2} (bottom, right): \textit{Smatch}, \textit{SRL} (Smatch computed on arg-i roles), \textit{IgnoreVars} (triple overlap after replacing variables with concepts) and \textit{Concepts} (F1 for concept identification).\footnote{\label{fn:tasks}The other subtasks are: \textit{Unlabelled} (Smatch after edge label removal), \textit{No WSD} (Smatch after PropBank sense removal), \textit{NS frames} (PropBank frame identification without sense), \textit{Wikification} (entity linking), \textit{NER} (named entity recognition),  \textit{Reentrancy} (Smatch over re-entrant edges).} GPLA produces the overall best parse but it is %system
%it
is outperformed
%excelled 
by %the parses of 
the other %two %automatic 
systems %F1-wise 
in
%sub-categories 
\textit{SRL} (JAMR) and \textit{IgnoreVars} (CAMR).
\begin{comment}
\begin{table}
  \centering
  \begin{tabular}{ll}
    \hline
    Metric & Description \\
    \hline
    Smatch F1 & triple matching with maximally scoring variable mapping\\
    Negation F1 & Smatch calculated with removed edge labels\\
    No WSD & not taking into account different sense (e.g.\ dive-01 equals dive-02)\\
    NP-only & Smatch on noun phrases only (see \cite{sawai2015semantic})\\
    Reentrancy & Reentrant nodes (co-reference, control)\\
    Concepts & Concept identification\\
    Named Ent. & NER (nodes with incoming :name edge)\\
    Wikification & Wikipedia KB linking (nodes with incoming :wiki edge)\\
    Negations & detection of negated concepts (in scope of :polarity -)\\
    SRL & semantic role labeling\\
    \hline
  \end{tabular}
  \caption{Evaluation of the two parses in Figure~\ref{fig:2parses} with the proposed evaluation suite.}
  \label{tab:metrics}
\end{table}
\end{comment}

\paragraph{Task definition} We 
%will 
adopt the proposed metrics by \citet{DBLP:journals/corr/DamonteCS16} and use them as target metrics for our task of AMR parse accuracy prediction.
Given 
an automatic AMR graph $G$ and a corresponding sentence $S$, we estimate precision, recall and F1 of the main task (Smatch)
%precision, recall and F1) 
\textit{and} of the subtasks, as they would emerge from comparing $G$ with its gold counterpart $G'$.

One of our hypotheses is that predicting a wide range of accuracy metric scores for individual aspects of AMR structures will aid our model to better predict the global Smatch scores. We will therefore investigate a hierarchical model that builds on predicted subtask measures in order to predict the global smatch score. Being able to predict fine-grained quality aspects of AMR parses will also be useful to assess and exploit differences of alternative system outputs and provides a basis for guiding system development or targeted annotation in an active learning setting.
%Robustness is less straightforwardly to define and we define it here as the ability of the estimation model to be invariant to different random initializations and to generalize well onto unseen domains.

\section{Neural Accuracy Prediction Model}% for AMR Parsing}
%\comkcomment{Neural Multi-Metrics Accuracy Prediction for AMR Parsing\\
%Neural AMR Accuracy Prediction with Multiple Metrics (MM-AMR)}\\
%A multi-metric regression model
%}}
\label{sec:model}

\begin{figure}
    \centering
    \includegraphics[scale=0.4]{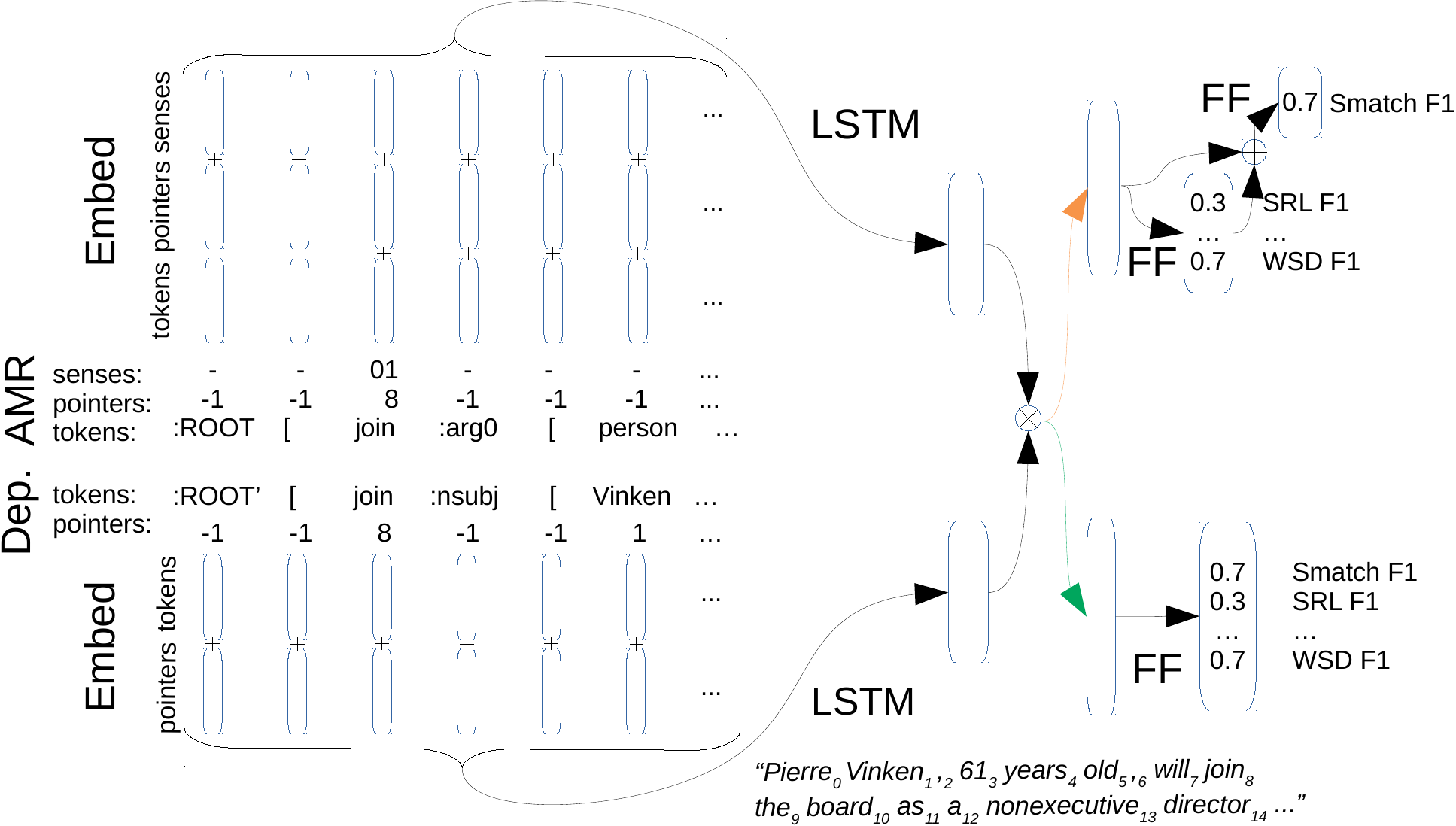}
    \caption{Our model:
    %outline. 
    green: 
    %All 
    Evaluation metrics 
    %are 
    computed in a non-hierarchical fashion. orange: Main evaluation metric is computed on top of secondary 
    %evaluation 
    metrics.}% losses are weighted.}
    \label{fig:model}
\end{figure}

%\comk{We propose a model for accuracy prediction for AMR parses that predicts multiple quality metrics, using a neural regression model.} 
We propose a neural hierarchical multi-output regression model for accuracy prediction of AMR parses. % that predicts multiple accuracy metrics, using a neural regression model.
%\comk{The architecture of our} 
Its architecture is 
outlined 
in Figure \ref{fig:model}. 
\paragraph{Inputs} Our model takes the following inputs: (i) a linearized AMR and a linearized dependency graph (implementation details in \S \ref{para:prepro}). The motivation for feeding the dependency parse instead of the original sentence is due to the moderate similarity of dependency and AMR structures.\footnote{c.f.\ \citet{DBLP:journals/corr/abs-1805-11465,E17-1053}.} We examine drawbacks and benefits of providing automatic dependency parses more closely in our ablation experiments (\S \ref{subsec:ablations}). %\footnote{We follow \citet{DBLP:journals/corr/KonstasIYCZ17}, \citet{DBLP:journals/corr/NoordB17a} in performing a depth-first traversal over the AMR to produce a sequence of node and edge labels, with brackets indicating the hierarchical structure. The same procedure is applied to the dependency parse of the input sentence.}
In addition, (ii)  we produce alignments between sentence tokens and tokens in the sequential AMR structure, as well as between sentence tokens and the linearized dependency structure, and feed these sequences of pointers to our accuracy prediction model. The intuition of using pointers is to provide the model with richer information via shallow alignment between AMR, dependencies and the sequence of sentence tokens (see Section \S \ref{para:prepro} for implementation details). Finally, (iii) we feed a sequence of PropBank sense indicators for AMR predicates. 

\paragraph{Joint encoding of AMR and dependency parses for metric prediction} Embedding layers are shared between AMR/dependency pointers and AMR/dependency tokens. We embed the three sequences representing the AMR graph (tokens, pointers and senses) in three matrices and sum them up element-wise
(indicated with $+$ in Figure \ref{fig:model}). The same procedure is applied to
%same as 
the 
%two matrices 
%of embedding sequences for the 
linearized dependency graph (tokens and pointers). 
The resulting matrices are processed by two two-layered Bi-LSTMs to yield vectorized representations for (i) the AMR graph and (ii) the dependency tree (i.e., the last states of forward and backward reads are concatenated). Thereafter, we apply element-wise multiplication, subtraction and addition to both vector representations and concatenate the resulting vectors ($\otimes$ in Figure \ref{fig:model}). The joint AMR-dependency 
representation is further processed by a %fully connected feed 
feed forward layer (FF) with 
%a sigmoidal 
sigmoid activation functions in order to predict, in total, 36 different metrics (\color{cadmiumgreen}green\color{black}, Figure \ref{fig:model}). 

\paragraph{Hierarchical prediction of multiple metrics} The task 
%also 
naturally lends itself to be formulated in a hierarchical multi-task setup (\color{orange}orange\color{black}, Figure \ref{fig:model}). In this strand, we first compute the 33 
%instead of 36 metrics 
 fine-grained subtask metrics %(for the subtasks of negation, WSD, SRL, etc.) 
and on their basis we caclulate the Smatch scores  (precision, recall, F1) as our primary metrics. In order to accomplish this, we collect the outputs from the subtask metric prediction layer in a vector and concatenate it with the previous layer's representation ($\oplus$ in Figure \ref{fig:model}). The resulting vector is fed through a last FF layer to predict the metrics for the task of main interest (Smatch). Our intuition is that the estimated quality of the parse 
%with respect to 
with respect to the subtask metrics informs the model and allows it to better predict
%on 
the overall quality. 
\paragraph{Loss} \label{par:loss}In the non-hierarchical case, we denote our full model with $f_\theta:X\rightarrow [0,1]^d$ 
%our full model with 
with parameters $\theta$, where $d$ describes the dimensionality of the score vector (one dimension represents one metric) and  $D=\{(X_i,y_i)\}_{i=1}^N, y_i \in [0,1]^d$ is our training data. 
%Then, 
In the non-hierarchical model, we minimize the mean squared error:
\begin{equation}
    \ell(f_\theta) =  \frac{1}{dN}\sum_{i=1}^N\sum_{j=1}^d (y_{i,j}-f_\theta(X_i)_j)^2
    \label{eq:nhl}
\end{equation}
For our hierarchical model, we have two functions, $f_{\theta}:X\rightarrow [0,1]^{(d-k)}$ which returns the output vector for the $(d-k)$ subtask metrics and $f'_{\theta'}:X\rightarrow [0,1]^k$ which returns the output vector for our $k$ main metrics (in our experiments, $k=3$ for Smatch recall, precision and F1). Then,
\begin{displaymath}
    \begin{split}
    \ell'(f_{\theta},f'_{\theta'}) =  \frac{\lambda_1}{(d-k)N}\sum_{i=1}^N\sum_{j=1}^{d-k} (y_{i,j}-f_{\theta}(X_i)_j)^2\\
    + \frac{\lambda_2}{kN}\sum_{i=1}^N\sum_{j=d-k+1}^d (y_{i,j}-f'_{\theta'}(X_i)_{j-(d-k)})^2
    \end{split}
\end{displaymath}
defines the total loss over the two entangled metric prediction models. Note that $\theta \subset \theta'$, 
which means that by optimizing the parameters of $f'$ with gradient descent% (second part of the sum)
, we also concurrently optimize all parameters of $f$. % (first part of the sum). %\comk{By this construction, the hierarchical model instantiates a multi-task model with shared parameters.} 
By this construction, the hierarchical model instantiates a two-task model with shared parameters. For our experiments, we manually set the loss weights $\lambda_1=0.2,\lambda_2=1$.
\section{Experiments}
\label{sec:experiments}

\paragraph{Data} Since our goal is to predict the accuracy of
%for 
an automatic parse, we need 
%to collect 
a data set containing 
automatically produced AMR
%deficient 
parses and 
their
%the 
scores, 
%of the different metrics of the automatic parses 
as they would emerge from comparison to gold parses. Our largest data set, LDC2015E86, comprises 19,572 sentences and comes in a predefined training, development and test split. We parse this data set with three parsers, JAMR \cite{P14-1134,DBLP:conf/semeval/FlaniganDSC16}, CAMR \cite{wang-xue-pradhan:2015:NAACL-HLT,wang-xue-pradhan:2015:ACL-IJCNLP,wang-EtAl:2016:SemEval} and 
GPLA
%GPLA 
\cite{DBLP:journals/corr/abs-1805-05286}. Since the three parsers have been trained on the training data partition, we naturally obtain more accurate parses for the training partition than for development and test data. 
\begin{table}

    \centering
    \scalebox{0.85}{\begin{tabular}{lrr|rr}
    \toprule
    &\multicolumn{2}{c|}{training}&\multicolumn{2}{c}{development}\\ \cmidrule{1-3}\cmidrule{4-5}
        parser & Smatch (F1) & \% def. & Smatch (F1) & \% def.\\
\midrule
        JAMR & 0.79 & 86.7 & 0.69 & 91.8 \\
        CAMR & 0.75 & 93.6 & 0.66 & 95.7\\
        GPLA & 0.86 & 83.4 &  0.76 & 90.0\\
        \bottomrule
    \end{tabular}}
    \caption{Parser output %vs.\ gold evaluation statistics 
    evaluation on training and
    development partitions of LDC2015E86. Smatch F1: avg. %average 
    over Smatch F1
    per sentence, \% def.: percentage of deficient parses (i.e., parses with Smatch F1 
   $<$
    %lower than
    1).}
    \label{tab:parser-stats}
\end{table}
Table \ref{tab:parser-stats}, however, indicates that we still 
%were able to 
obtain a considerable amount of deficient parses for training. Based on the parser outputs we compute evaluations comparing the automatic  
parses with the gold parses by using \textit{amr-evaluation-tool-enhanced}\footnote{\url{https://github.com/ChunchuanLv/amr-evaluation-tool-enhanced}}, a bug-fixed version of the script 
%for computation of 
that computes
the metrics of
%introduced by 
\citet{DBLP:journals/corr/DamonteCS16}. This 
%step 
allows us to create full-fledged training, development and test instances
%examples 
for our accuracy prediction task. Each 
%data example
instance consists of a sentence and an 
%an 
AMR parse 
%and sentence 
as input and a vector of metric scores as target. %The target vector has 36 dimensions (Smatch and eleven sub tasks with three metrics each: precision, recall, F1)
%different 
%metrics \comk{with}
%precision, recall and F1 
%for 
%every one of the 
%). 

Our second data set, LDC2015R36, comprises submissions to the SemEval-2016 Task 8 \cite{may:2016:SemEval}.
%in other words, 
We have 1053 parses from each of the 11 team submissions (and 2 baseline systems).\footnote{\label{fn:teams}Riga \cite{barzdins2016riga}, CMU (equal to 
JAMR) \cite{DBLP:conf/semeval/FlaniganDSC16}, Brandeis \cite{wang2016CAMR}, UofR \cite{peng2016uofr}, ICL-HD \cite{S16-1179}, M2L \cite{DBLP:conf/semeval/PuzikovKK16}, UMD \cite{S16-1184}, DynamicPower \cite{S16-1177}, TMF \cite{S16-1182}, UCL+Sheffield \cite{S16-1180} and CU-NLP \cite{S16-1185}.} Our third dataset, BioAMRTest is used as the test set in the SemEval-2017 Task 9 \cite{may-priyadarshi:2017:SemEval} and consists of 500 parses from each of the 6 teams.\footnote{\label{fn:teamsbio}TMF-1 and TMF-2 \cite{S17-2160}, DANGNT \cite{DBLP:conf/semeval/NguyenN17}, Oxford \cite{11100}, RIGOTRIO \cite{S17-2159} and JAMR \cite{DBLP:conf/semeval/FlaniganDSC16}} The shared task organizers kindly made this data available for our experiments.
\begin{table}%[]
%\scriptsize
\scalebox{0.72}{
    \centering
    \begin{tabular}{llllll}
         \toprule
         data set & LDC2015E86 & LDC2015R36 & BioAMRTest  \\
         \midrule
         domain & news & news & medical  \\
         %domain & news & 19,572 & &58,716 & \\
         nb.\ sentences & 19,572 & 1,053 & 500 \\
         avg.\ sent.\ len.\ & 21 & 22.35 & 36.52 \\
         nb.\ auto.\ parses & 58,716 & 13,689 & 3,000 \\
         %automatic parses & medical & 500 & &3,000 \\
         
         used as & train/dev/test&  test &test \\
         %mf tokens & and & and& ref\\
          %&not & his& cell\\
          %&from&he& Figure\\
          %&stated&I& type=fig\\ 
          %&they&they&i\\
          %&n't&would&cells\\ 
         % &would&were&phosphorylation\\
         % &China&He&xref \\
         % mf edges & :arg1 & :arg1 & :op1 \\
         % & :op1 &:op1&:arg1\\
         % & :arg0& :name&:name \\
        %  & :mod & :arg0& :mod\\
         % & :name & :mod& :arg1-of\\
        %  & :wiki & :op2& :arg0\\
        %  &:arg2 & :arg2& :arg2\\
        %  mf concepts & name & name & and \\
        %  & country &person& protein\\
        %  & and& and&enzyme \\
        %  & person& country& cell-line\\
        %  & I & say-01& phosphorylate-01\\
        %  & date-entity & organization& describe-01\\
        %  &have-org-role-91 & date-entity& mutate-01\\
         \bottomrule
         %mf concepts & train/dev/test&  test &test \\
         %mf edges & train/dev/test &  test &test \\
    \end{tabular}}
    \caption{Statistics of data sets used in this work. 
    }
    \label{tab:datasets}
\end{table}
\paragraph{Preprocessing}\label{para:prepro} For dependency annotation, we parse all sentences with spacyV2.0\footnote{https://spacy.io/}. %, a multi-task CNN based parser trained on OntoNotes, with GloVe vectors trained on Common Crawl. 
For sequentializing the AMR and dependency graph representations we take intuitions from \citet{DBLP:journals/corr/NoordB17a} \& \citet{DBLP:journals/corr/KonstasIYCZ17} and 
%gather the tokens via 
output tokens by performing a depth-first-search over the graph. 
%where the 
%variables have been replaces by their concepts. 
We replace the AMR negation token `-' and 
%also 
strings representing numbers with special tokens. The vocabularies (tokens, senses and pointers) are computed from our training partition of LDC2015E86 and comprise all tokens with a frequency $\geq$ 5 (tokens with lesser frequency are replaced by an \textit{OOV}-token). 
PropBank senses of predicates are removed and collected in an extra list that is parallel to the tokens in the linearized AMR sequence. For each linearized AMR and dependency tree
%sequence 
we generate a sequence with index pointers to tokens in the original sentence (-1 for tokens which do not explicitly refer to any token in the sentence, e.g.\ brackets, `subj' or `arg0' relations). Extraction 
of token-pointers from
%from 
the dependency graph is trivial. %because it is an annotation over the words in the sentence. 
For every concept in the linearized AMR we execute a %simple 
search for the corresponding token in the sentence, looking for exact matches with  surface tokens and lemmas. 

\paragraph{Training} For the optimization of the accuracy prediction model we use only the development and training sections of LDC2015E86 and the corresponding automatic parses together with the gold scores. Details on the training cycle can be found in the \textit{Supplemental Material} \S \ref{sec:supplemental} (the loss is described in \S \ref{par:loss}). We use the same single (hierarchical) model for all three evaluation studies, proving its applicability across different scenarios (a non-hierarchical model is only instantiated for the ablation experiments in Section \S \ref{subsec:ablations}).

\subsection{Correlation with Gold Accuracy}

The primary goal in our first
%this 
experiment is to test whether the system is able to differentiate good from bad parses. This capacity is expressed by a high correlation of predicted accuracies with true accuracies on unseen data and by the ability to assign high scores to gold parses. We evaluate on the test partition of LDC2015E86 and BioAMRTest. 

\paragraph{Correlation results} The results are displayed in Table \ref{tab:mainres}.
\begin{table}
\begin{center}
\scalebox{0.87}{
\begin{tabular}{@{}lrrr|rrr@{}} 
%& & & & cost & &&time (ms)  \\
&\multicolumn{3}{c|}{$\rho$ LDC2015E86}&\multicolumn{3}{c}{$\rho$ BioAMRTest}\\ \cmidrule{2-4}\cmidrule{5-7}
%\multicolumn{11}{l|}{...}\\
 & P & R & F1 & P & R & F1 \\%\\
\toprule
 Smatch &0.74 & 0.79 & 0.78 & 0.54 & 0.41 & 0.47 \\
\midrule
Concepts  & 0.56 & 0.65 & 0.64 & 0.67 & 0.55 & 0.62 \\
Frames  & 0.7 & 0.71 & 0.72 & 0.67 & 0.56 & 0.63\\
IgnoreVars  & 0.76 & 0.8 & 0.79 & 0.33 & 0.27 & 0.29\\
Named Ent.  & 0.81 & 0.81 & 0.81 & 0.5 & 0.48 & 0.5 \\
Negations  & 0.87 & 0.87 & 0.87 & 0.33 & 0.32 & 0.32\\
No WSD  & 0.75 & 0.78 & 0.78 & 0.54 & 0.41 & 0.46\\
NS frames  & 0.76 & 0.75 & 0.77 & 0.72 & 0.59 & 0.67\\
Reentrancies  & 0.77 & 0.79 & 0.8 & 0.52 & 0.45 & 0.48\\
SRL  & 0.72 & 0.74 & 0.75 & 0.47 & 0.43 & 0.45\\
Unlabeled  & 0.71 & 0.75 & 0.75 & 0.6 & 0.45 & 0.51\\
Wikification  & 0.87 & 0.85 & 0.86 & 0.24 & 0.23 & 0.23 \\
%Concepts  & 0.57 & 0.64 & 0.64 & 0.57 & 0.51 & 0.56 \\
%Frames  & 0.69 & 0.71 & 0.72 & 0.63 & 0.51 & 0.57 \\
%IgnoreVars & 0.77 & 0.81 & 0.8 & 0.29 & 0.28 & 0.29 \\
%NER & 0.82 & 0.81 & 0.82  & 0.43 & 0.40 & 0.42 \\
%Negations  & 0.88 & 0.88 & 0.88  & 0.32 & 0.31 & 0.32 \\
% No WSD  & 0.75 & 0.78 & 0.78 & 0.45 & 0.38 & 0.41 \\
%NS-frames  & 0.75 & 0.75 & 0.77  & 0.70 & 0.54 & 0.63 \\
%Reentrancies  & 0.76 & 0.78 & 0.79  & 0.43 & 0.40 & 0.42 \\
%SRL  & 0.71 & 0.75 & 0.75 & 0.44 & 0.40 & 0.42 \\
%Unlabeled  & 0.71 & 0.76 & 0.76& 0.52 & 0.42 & 0.47 \\
%Wikification  & 0.82 & 0.79 & 0.81 & 0.39 & 0.38 & 0.38 \\
\bottomrule
\end{tabular}}
\end{center}
\caption{Pearson correlation coefficient ($\rho$) over various metrics and across domains. Explanations of the metrics and AMR subtasks are in Section \S \ref{sec:tasks} and fn.\ \ref{fn:tasks}}
\label{tab:mainres}
\end{table}
\begin{figure}%[h]
\subcaptionbox{LDC2015E86 (train)\label{fig3:a}}{\includegraphics[width=1.5in]{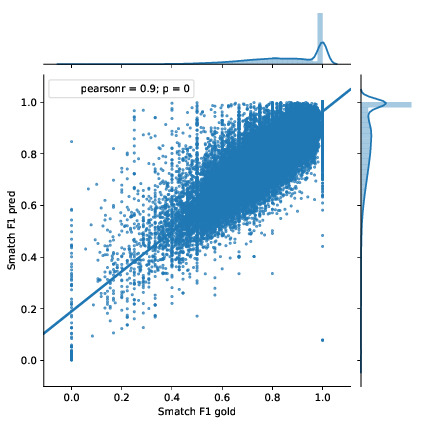}}%\hspace{1em}%
  \subcaptionbox{LDC2015E86 (test)\label{fig3:b}}{\includegraphics[width=1.5in]{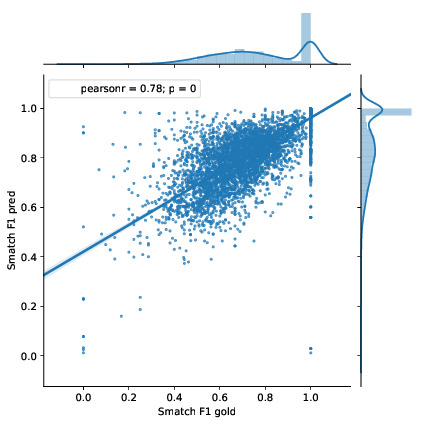}}%\hspace{1em}
  %\subcaptionbox{3a\label{fig3:a}}{\includegraphics[width=1.42in]{Smatchf1goldLDC.pdf}}\hspace{1em}%
  %\subcaptionbox{3b\label{fig3:b}}{\includegraphics[width=1.42in]{Smatchf1goldBIO.pdf}}
  \caption{\hspace*{-1mm}Predicted (y-axis) \& gold (x-axis) Smatch F1.}
  \label{fig:corr}
\end{figure}
Over all metrics, in-domain and out-of-domain, we achieve significant correlations  with the gold scores ($p<0.005$ for every metric). While on LDC2015E86 the model has learned to predict the KB linking F1 ($\rho = 0.86$) and negation detection F1 with high correlation to the gold scores ($\rho =0.87$), Concept assessment poses the greatest challenge ($\rho = 0.64$). For the out-of-domain data BioAMRTest, these two facts seem almost 
%seem 
reversed: here, the assessment of 
%the 
KB linking poses difficulties ($\rho=0.23$) while the Concept F1 predictions are better ($\rho =0.62$). The main metrics of interest (Smatch precision, recall and F1) can be predicted with high correlation on in-domain data ($\rho\geq 0.74$, cf.\ also Figure \ref{fig:corr}) and solid correlation for out-of-domain data ($\rho \geq 0.41$).

\begin{figure}
\centering
\begin{subfigure}[b]{0.5\textwidth}
   \includegraphics[width=1\linewidth]{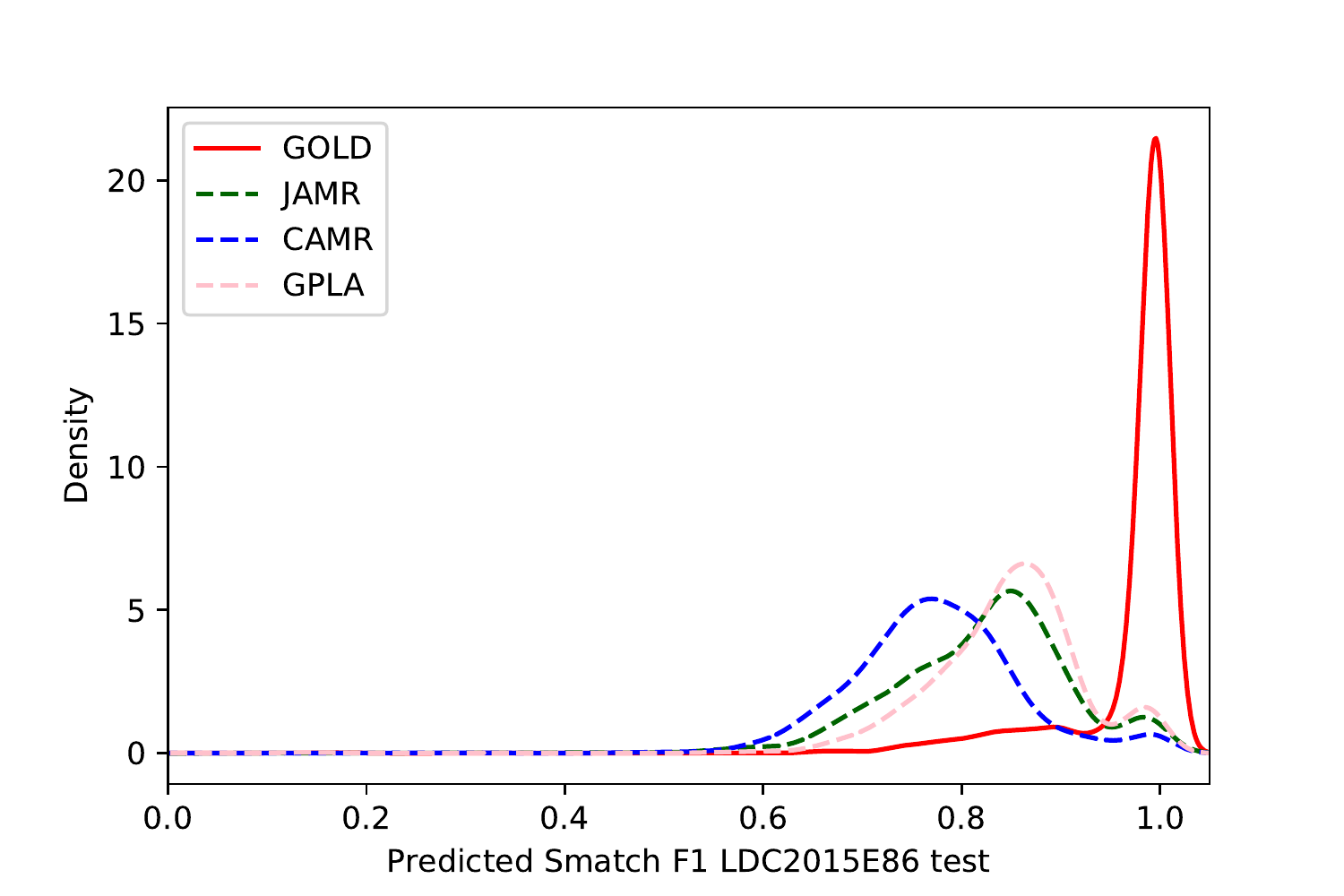}
   \caption{LDC2015E86}
   \label{fig:Ng1} 
\end{subfigure}

\begin{subfigure}[b]{0.5\textwidth}
   \includegraphics[width=1\linewidth]{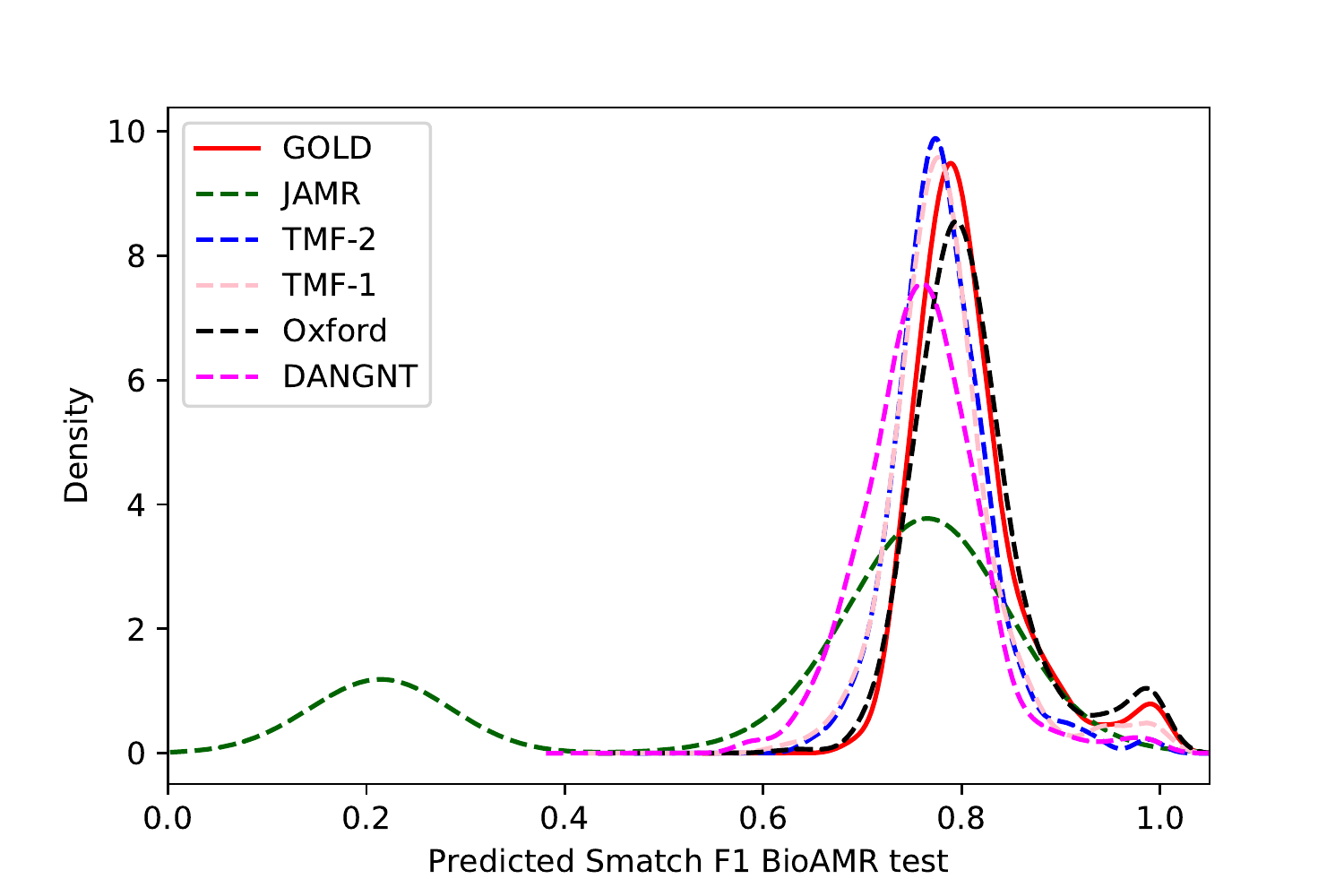}
   \caption{BioAMRTest}
   \label{fig:Ng2}
\end{subfigure}

\caption{Probability density function estimations for predicted F1 Smatch scores using Scott's method  \cite{scott2012multivariate} with respect to candidate parses from different systems.}
\end{figure}

\paragraph{Find the Gold AMR!} Now, we want to test our system's capacity to reliably predict high Smatch F1 scores for unseen gold AMR parses. Ideally, the scores should be close or equal to 1. For in-domain data, it appears to work well: a large amount of Smatch predictions for gold AMR graphs are very close to one (Figure \ref{fig:Ng1}).

\begin{table}
    \centering
    \scalebox{0.77}{
    \begin{tabular}{lrrrrrrrr}
    &\multicolumn{7}{c}{percentile}\\
    \toprule
    
   dataset &5 & 25 & 75&90&95&97&99 \\
    \midrule
        LDC15E86 &0.83&0.99&1.0&1.0&1.0&1.0&1.0 \\
        BioAMRTest &0.74&0.77&0.83&0.88&0.93&0.98&1.0 \\
        \bottomrule
    \end{tabular}}
    \caption{Various percentiles of Smatch F1 predictions for gold graphs.}
    \label{tab:percentilesgold}
\end{table}
Evidently, our system also gets the ranking of the parsing systems right: the distribution of the state-of-the-art (GPLA) is shifted right towards higher predicted F1 scores, whereas the distribution of CAMR is shifted left towards lower scores. Also, more than 75\% of gold parses have a predicted Smatch score of more than 0.99 (Table \ref{tab:percentilesgold}).

On the other hand, finding gold parses in the BioAMRtest data is much harder: about 75\% of Smatch scores get assigned a score of 0.83 or lower and only 1\% of gold parses are predicted as perfect (Table \ref{tab:percentilesgold}). The estimated probability density function for gold parses (red solid line in Figure \ref{fig:Ng2}) struggles to discriminate itself from the functions corresponding to the flawed parses of the automatic systems. Nevertheless, the prediction score density for gold parses is situated more on the right hand side than most others. In other words, we find that in the out-of-domain data gold parses tend to be assigned above-average scores. 

To sum up, our observations for the out-of-domain data stand in some contrast to what we observe for the in-domain data. However, this outcome can be plausibly explained: assuming that the out-of-domain gold parses have some unfamiliar properties, a system that has never seen such parses cannot judge well whether they are gold or not. In fact, it can be interpreted positively that the system hesitates to assign maximum scores to gold parses from a domain in which the model is completely inexperienced. Additionally, bio-medial texts involve difficult concepts, naming conventions and complicated noun phrases which are hard to understand even for non-expert humans (e.g., \textit{``TAK733 led to a decrease in pERK and G1 arrest in most of these melanoma cell lines regardless of their origin, driver oncogenic mutations and in vitro sensitivity to TAK733''.}). Taking all this into account, the results for out-of-domain data may be not as bad as they perhaps appear at first glance.

\subsection{Application Study: AMR Parse Ranking}
\label{subsec:app1}
Our automatic accuracy prediction method naturally lends itself for ranking parser outputs. For any sentence, 
%several 
provided automatic 
%candidate 
parses by competing systems can be ranked according to the scores predicted by our system. This scenario arises, e.g., when we run several AMR parsers over a large corpus with the aim of selecting the
%and needs to determine the 
best parse for each sentence in order to collect silver training data.\footnote{In a self-training scenario, we also could set a threshold of minimum predicted accuracy to select confident parses.} In the worst case, we do not have any prior knowledge about a parser's performance (we may not even know the source of a parse). %\footnote{E.g., the parser is applied to a completely different domain}. 
We use the test partition from LDC2015E86 and BioAMRTest to rank, for each sentence, the automatic candidate parses provided by the different parsers. In LDC2015E86 we assume not to be agnostic about the parsers as their performances on the development data 
%are
of this data set are known (in terms of their sentence-average F1 Smatch score). 
%For each parser we know its average Smatch F1 on the development data. 
%This means, 
Consider that we are
given a sentence and three 
%alternative 
automatic parses. We select the maximum-score parse, 
%best-parse prediction 
 where the score is defined by predicted Smatch F1 plus the average Smatch F1 of the parse-producing parser on the development data. 
As baselines
%The baseline 
in this scenario 
we 
%is
(i) randomly choose
%ing 
a parse from the three options  
or (ii) always choose
%ing 
the parse of GPLA. On BioAMRTest, however, we have no prior information about the submitted systems. We %can decide between
select from 6 
%different 
automatic parses for 
each
%every 
sentence.  
Since now
%Because 
we are completely parser agnostic,
%team's 
the 
%only 
baseline is
%, for every sentence, 
to 
%uniformly 
randomly select a parse from the candidate set.
%select a parse 
%from any of the teams' \comk{parses}.
%[TODO: Discuss rsults]
\paragraph{Results} 
\begin{table}
%\scriptsize
\scalebox{0.77}{
\begin{tabular}{@{}lrrr|rrr|} 
%& & & & cost & &&time (ms)  \\
&\multicolumn{3}{c|}{Smatch LDC2015E86}&\multicolumn{3}{c|}{Smatch BioAMRTest}\\ \cmidrule{2-4}\cmidrule{5-7}
%\multicolumn{11}{l|}{...}\\
 & P & R & F1 & P & R & F1\\ %\\
\toprule
lower-bound &  64.9 & 57.9 &60.5  & 41.7 & 31.3&34.3 \\
random &  72.4 & 67.0  &69.1 &  60.3 &  50.3& 54.0 \\
ours &\underline{\textbf{76.6}} & \underline{\textbf{73.5}}& \underline{\textbf{74.8}}& \underline{64.9} &\underline{56.0} &\underline{59.2}\\
upper-bound  &  79.3 & 75.2 &76.9 & 73.2 & 65.2& 68.5\\
\midrule
JAMR & 71.4 & 66.5& 68.4 &  48.4 & 39.7 & 42.9 \\
CAMR &  69.5 & 60.4 &64.0&-&-&-\\
GPLA & 76.3 & 73.4 & 74.6 &- &- & -\\
TMF-1 & - & - & - & 56.0&46.5 &49.3\\
TMF-2 & - &  - &  - &  70.0 & 54.5& 60.5\\
DANGNT & - &  - & -   & \textbf{70.2} & 58.6 & \textbf{63.1} \\
Oxford & - & - & - &  65.8 & \textbf{59.0}  & 61.6 \\
RIGOTRIO & - & - & -   &  65.0 &  50.8 & 56.4\\
\bottomrule

\end{tabular}}
\caption{Results (sentence averages) of different AMR parsing (bottom part) and ranking (top part) systems on two test sets. Upper part: results when selecting from alternative parses: lower-bound (upper-bound): oracle 
%which 
%always 
selecting the worst (best)
%out of various 
AMR parse; % for a sentence;
ours: results when selecting the best parse according to our models' accuracy prediction (hierarchical model). }
\label{tab:res-smatch}
\end{table}
The results are displayed in Table \ref{tab:res-smatch}. For our in-domain test data,
LDC2015E86, selecting the best parse according to our model's predicted accuracy score improves over \textit{all} individual parser results: 
%it \textit{always} helps to rank the different automatic parses on a sentence level using our model: 
the obtained average Smatch F1 per sentence 
%is 
increases (i) slightly by 0.2 pp.\ compared to always choosing outputs from GPLA and (ii) observably by 5.7 pp.\ compared to randomly selecting a parse from the competing system outputs. 
%On the one hand,
The difference compared to always choosing GPLA seems negligible which perhaps can be explained by the fact that GPLA has been shown to be on par or better than doubly-blind human annotators.\footnote{GPLA \cite{DBLP:journals/corr/abs-1805-05286} 
%has been shown to 
achieves a high
%as high as 
74.4\% cor\-pus-\-lev\-el Smatch F1 (primarily news texts), while a prior annotation study \cite{Banarescu13abstractmeaning} reported doubly blind annotation corpus-level F1 of 0.71 (for web texts).}
%On the other hand, 
%Thus, this is a most challenging and perhaps even impossible task. 
%An all-knowing 
The oracle 
%who 
that always selects the best parse (\textit{upper-bound} in Table \ref{tab:res-smatch}) shows little room for improvement: it achieves
%is by 
2.1 pp.\ Smatch F1 
increase compared to
%better than 
our model. 
%While the upper bound 
%shows 
%indicates that there is some room for improvement, 
This margin is small and further success might also be hampered by peculiarities in the manual annotations.  On BioAMRTest, no prior information about the systems is available.
Using our model's predicted scores to select from the alternative system outputs, %Applying our system to select for every sentence the best parse, 
we can boost Smatch F1 by 5.2 pp.\ compared to randomly selecting a parse. Compared to always selecting the parses of the best submitted system (in-hindsight), we lag behind by 3.9 pp.

\begin{table}
\scalebox{.8}{
\begin{tabular}{@{}lrr|rr@{}} 
%& & & & cost & &&time (ms)  \\
&\multicolumn{2}{c|}{Smatch LDC2015E86}&\multicolumn{2}{c@{}}{Smatch  BioAMRTest}\\ \cmidrule{2-3}\cmidrule{4-5}
%\multicolumn{11}{l|}{...}\\
 & $\bar{\rho}$ & \%pos & $\bar{\rho}$ & \%pos\\   %\\
\toprule
lower-bound &  -1 & 0.0 & -1  & 0.0 \\
random &  0.00 & 50.0  &0.00 &  50.0   \\
ours & \textbf{0.54} & \textbf{77.0} &\textbf{0.22}&\textbf{70.4} \\
%dev based & \textbf{0.54} & \textbf{77.0} &
%-&- \\
upper-bound  &  1.00 & 100.0 &1.00 & 100.0  \\
%\midrule
%JAMR & 0.714 & 0.665& 0.684 &  0.484 & 0.397 & 0.429 \\
%CAMR &  0.695 & 0.604 &0.640&-&-&-\\
%GPLA & 0.763 & 0.734 & 0.746 &- &- & -\\
%TMF-1 & - & - & - & 0.560&0.465 &0.493\\
%TMF-2 & - &  - &  - &  0.700 & 0.545& 0.605\\
%DANGNT & - &  - & -   & \textbf{0.702} & 0.586 & \textbf{0.631} \\
%Oxford & - & - & - &  0.658 & \textbf{0.590}  & 0.616 \\
%RIGOTRIO & - & - & -   &  0.650 &  0.508 & 0.564\\
%\midrule
%Concepts  & 0.57 & 0.64 & 0.64 & 0.57 & 0.51 & 0.56 \\
%Frames  & 0.69 & 0.71 & 0.72 & 0.63 & 0.51 & 0.57 \\
%IgnoreVars & 0.77 & 0.81 & 0.8 & 0.29 & 0.28 & 0.29 \\
%NER & 0.82 & 0.81 & 0.82  & 0.43 & 0.40 & 0.42 \\
%Negations  & 0.88 & 0.88 & 0.88  & 0.32 & 0.31 & 0.32 \\
% No WSD  & 0.75 & 0.78 & 0.78 & 0.45 & 0.38 & 0.41 \\
%NS-frames  & 0.75 & 0.75 & 0.77  & 0.70 & 0.54 & 0.63 \\
%Reentrancies  & 0.76 & 0.78 & 0.79  & 0.43 & 0.40 & 0.42 \\
%SRL  & 0.71 & 0.75 & 0.75 & 0.44 & 0.40 & 0.42 \\
%Unlabeled  & 0.71 & 0.76 & 0.76& 0.52 & 0.42 & 0.47 \\
%Wikification  & 0.82 & 0.79 & 0.81 & 0.39 & 0.38 & 0.38 \\

\bottomrule

\end{tabular}}
\caption{Results of different parse-ranking systems with respect to sentence-level parse rankings. $\bar{\rho}$: average Pearson-r on a sentence level. \%pos: ratio of predicted rankings with positive $\rho$ to gold ranking. }
\label{tab:parsers-rankings}
\end{table}

\begin{figure}
    \centering
    \includegraphics[scale=0.4]{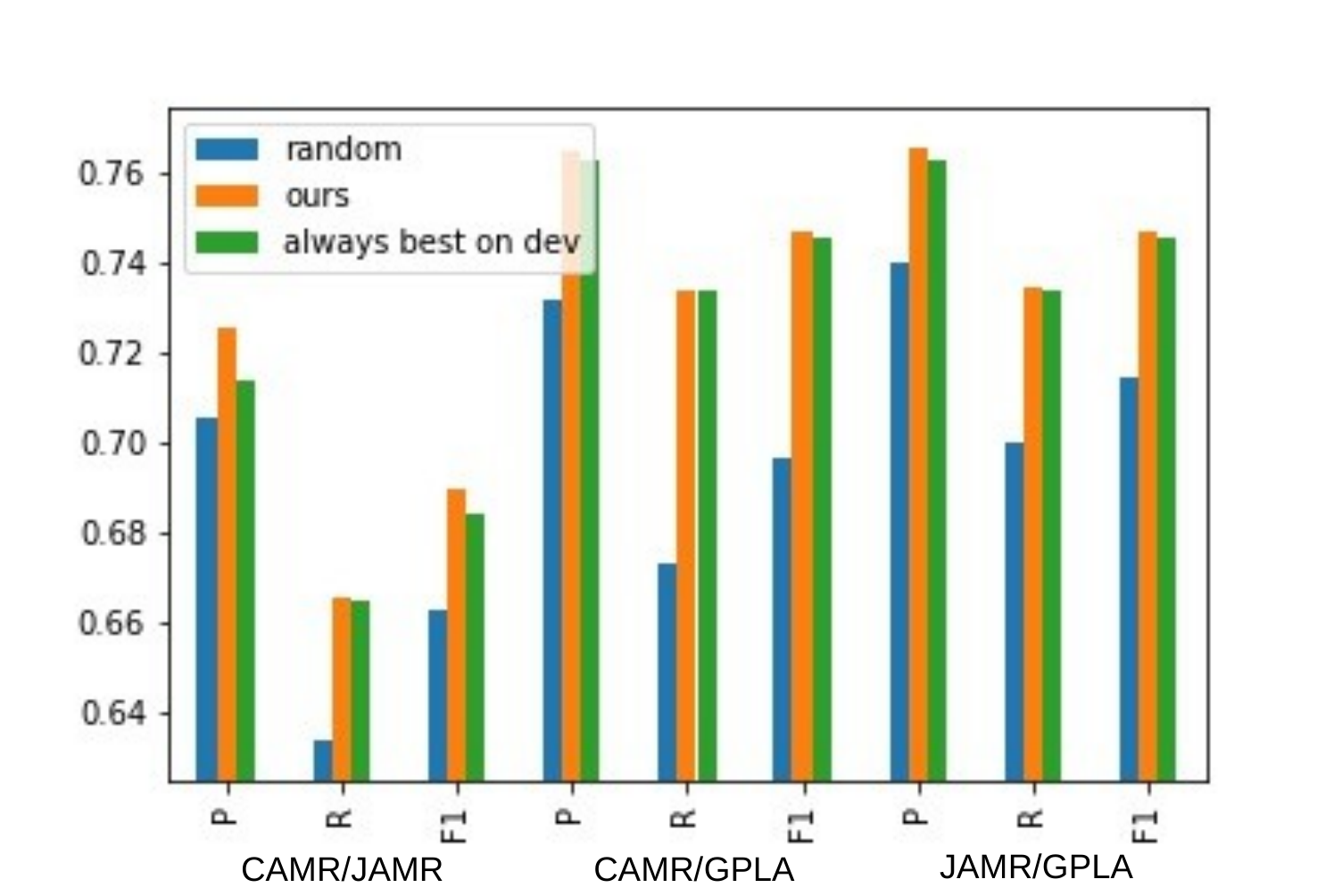}
    \caption{Using our model to predict the best parse out of two candidate parses, each from a different system.}
%    Using our system to select from two candidate parses for each sentence, each from a different system.}
    \label{fig:select_from}
\end{figure}

Since 
%in 
our 
%testing  
data comprises
%are 
outputs from 
several parsers with varying performance, we can study the performance of our approach in combination with different parsers %with respect to different parsers 
(Figure \ref{fig:select_from}). When only choosing among
%we can only chose from 
CAMR and JAMR outputs, on LDC2015E86, our system 
%is useful and 
boosts the F1 by 2.7 pp.\ compared to randomly selecting a parse, and 
%also 
by 0.6 pp.\ compared to always choosing 
%always 
the parse from
%by 
the better 
system
%parser 
(determined on dev, here: JAMR). Choosing from CAMR and GPLA or JAMR and GPLA makes little difference: in most cases our system selects the GPLA parse and the difference to only choosing GPLA parses is
%stays 
marginal. Moreover, across both test sets, the majority of rankings assigned by our method have positive correlations with the true rankings (Table \ref{tab:parsers-rankings}): 77\% of all assigned rankings  have a positive correlation with the true ranking (70\% for bio-medical).
In sum,
%summary,
%To summarize, from the experiments 
we can draw two conclusions from this experiment:
given a sentence, ranking 
AMR parser outputs using our accuracy prediction model, on in-domain and out-of-domain unseen data %different automatic parses with our model 
(i) clearly improves performance
%observably helps 
when 
non state-of-the-art parsers are applied or if we are not informed about the parsers' performances and (ii) does not worsen 
%the 
results in other cases. %Improvement over GPLA results are marginal, but may still be profitable in out-of domain contexts.

\subsection{Application Study: Predict System Ranks}
\label{subsec:app2}
%Finding the Best System}

In our final
%this 
case study, 
we 
%our aim is to 
use our accuracy prediction model %for 
%in order 
to predict a ranking over systems.
%-- without available gold data. 
%In our setup we use the model that
%Since our model 
%has been optimized on automatic parses from GPLA and JAMR and CAMR 
We use our model
%we
to rank the unseen submitted system parses %from 
of the SemEval-2017 Task 9 (evaluated on BioAMRTest) and SemEval-2016 Task 8 (evaluated on LDC2015R36) according to average predicted F1 Smatch scores. 
Again, we assume a parser-agnostic setting, meaning we have no prior knowledge of the submitted systems (i.e.\ we just consider their outputs). In this setting, 
%however, 
we do not rank individual parses given a sentence, but rank the system outputs, according to estimated average Smatch F1 per sentence. We evaluate against the final team rankings of the two shared tasks. 
\begin{table}
%\scriptsize
\begin{center}
\scalebox{.58}{
\begin{tabular}{@{}lrr|rr|@{}} 
%& & & & cost & &&time (ms)  \\
&\multicolumn{2}{c|}{Rank LDC2015R36}&\multicolumn{2}{c|}{Rank BioAMRTest}\\ %\cmidrule{2-3}\cmidrule{4-5}
%\multicolumn{11}{l|}{...}\\
 & rank $r$ & rank $\hat{r}$ &  rank $r$& rank $\hat{r}$ \\%\\
\toprule
DANGNT & -& -& 1 & 3   \\
Oxford & -& -& 2 & 1   \\
TMF-2 &- &- & 3 &  2  \\
RIGOTRIO &- &- & 4 & 5  \\
TMF-1 &- & -& 5 & 4  \\
JAMR &7 &7 &6 & 6  \\
RIGA & 1 & 4 & -&  - \\
Brandeis& 2& 3 &- &  - \\
CU-NLP& 3&1 &- &  - \\
UCL+Sheffield&4 & 2 & -& -  \\
ICL-HD&5 & 8 &- & -  \\
M2L& 6& 10& -&-   \\
%JAMR=CMU
JAMR-base& 8& 12 &- &  - \\
UofR& 9& 11& -& -  \\
TMF& 10& 5 &- &-   \\
UMD &11 & 6 & -& -  \\
DynamicPower& 12 & 13 & -& -  \\
det.\ baseline& 13 & 9 &- & -  \\
\midrule
$\rho$&\multicolumn{2}{c|}{0.645 ($p_1=0.017$, $p_2=0.011$)}&\multicolumn{2}{c|}{0.771 ($p_1=0.072$, $p_2=0.051$)}\\

%\midrule
%Concepts  & 0.57 & 0.64 & 0.64 & 0.57 & 0.51 & 0.56 \\
%Frames  & 0.69 & 0.71 & 0.72 & 0.63 & 0.51 & 0.57 \\
%IgnoreVars & 0.77 & 0.81 & 0.8 & 0.29 & 0.28 & 0.29 \\
%NER & 0.82 & 0.81 & 0.82  & 0.43 & 0.40 & 0.42 \\
%Negations  & 0.88 & 0.88 & 0.88  & 0.32 & 0.31 & 0.32 \\
% No WSD  & 0.75 & 0.78 & 0.78 & 0.45 & 0.38 & 0.41 \\
%NS-frames  & 0.75 & 0.75 & 0.77  & 0.70 & 0.54 & 0.63 \\
%Reentrancies  & 0.76 & 0.78 & 0.79  & 0.43 & 0.40 & 0.42 \\
%SRL  & 0.71 & 0.75 & 0.75 & 0.44 & 0.40 & 0.42 \\
%Unlabeled  & 0.71 & 0.76 & 0.76& 0.52 & 0.42 & 0.47 \\
%Wikification  & 0.82 & 0.79 & 0.81 & 0.39 & 0.38 & 0.38 \\

\bottomrule

\end{tabular}}
\end{center}
\caption{True rank $r$ 
(given corpus-Smatch) 
and predicted rank $\hat{r}$ (based on 
%as predicted by our system (according to 
sentence average Smatch computed using our model). $p_1$: probability of non-correlation. $p_2$: probability that a randomly produced ranking achieves equal or greater $\rho$ (estimated over $10^6$ random rankings). For team names, see fn.\ \ref{fn:teams} \& \ref{fn:teamsbio}.
%per sentence)
}
\label{tab:bestp}
\end{table}

\paragraph{Results} The results are displayed in Table \ref{tab:bestp}. On BioAMRTest we have a good, albeit non statistically significant correlation with the true team ranking. On the in-domain LDC2015R36 test set we see a significant correlation of $\rho=0.645$ ($p_{1,2}<0.05$). %\footnote{Note that teams were ranked according to true corpus F1, while we rank the teams according to averaged
%sentence averageSmatch F1 predicted for individual parses (since our method predicts Smatch not on a corpus level but on a sentence level).} 
In this shared task, many teams were competitive and differences between the best teams were marginal. For example, in the true ranking, places 1 to 6 achieved between 0.60 and 0.62 Smatch F1. Notably, the first four teams according to the true ranking and the first four teams according to our predicted ranking fall into the same group.
%are equal.
This shows that our model successfully assigned high ranks to low error submissions.

\subsection{Ablation Experiments}
\label{subsec:ablations}
We finally perform ablation experiments 
%in order 
to evaluate the impact of individual model components.
We experiment with five 
%different 
different setups. (i) instead of stacking two Bi-LSTMs, we use only one Bi-LSTM (\textit{one-lstm}, Table \ref{tab:ablations}). (ii) instead of the dependency tree, we feed the words in the order as they occur in the sentence (\textit{no-dep}). (iii) \textit{no-pointers}: we remove the token-pointers from our model. (iv), instead of using the hierarchical setup, we predict all metrics on the same level (\textcolor{cadmiumgreen}{green} in Figure \ref{fig:model}, \textit{no-HL} in Table \ref{tab:ablations}) and (v), \textit{no-HMTL}: we optimize the non-hierarchical model only with respect to Smatch, disregarding the AMR subtasks. %The results are displayed in Table \ref{tab:ablations}. 
Remarkably, the dependency tree greatly helps the model on in-domain data over all measures (-37 total $\Delta$ without dependencies) but hurts the model on out-of-domain data (+27 total $\Delta$). A possible explanation is the degradation of the dependency parse quality: bio-medical data not only poses a challenge for our model, but also for the dependency parser. With special regard to the main AMR evaluation measure, Smatch F1, the learned pointer embeddings provide useful input  on the in-domain test data (-4 $\Delta$ without pointers).

\begin{table}
\renewcommand*\rot[2]{\multicolumn{1}{R{#1}{#2}}}% no optional argument here, please!
%\scriptsize
\scalebox{.59}{
\begin{tabular}{l|r|rrrrr|r|rrrrr|}
%\scriptsize
&\multicolumn{6}{c|}{LDC2015R36}&\multicolumn{6}{c|}{BioAMRTest}\\

&
\rot{70}{1em}{complete} &
\rot{70}{1em}{one lstm} &
\rot{70}{1em}{no-dep} &
\rot{70}{1em}{no-pointers} &
\rot{70}{1em}{no-HL} &
\rot{70}{1em}{no-HMTL} &
\rot{70}{1em}{complete} &
\rot{70}{1em}{one lstm} &
\rot{70}{1em}{no-dep} &
\rot{70}{1em}{no-pointers} &
\rot{70}{1em}{no-HL} &
\rot{70}{1em}{no-HMTL} \\
\toprule
&$\rho$&\multicolumn{5}{c|}{$\Delta$}&$\rho$&\multicolumn{5}{c|}{$\Delta$}\\
\toprule
Smatch  & 78&-1&-1&-4 &  -3   &-2 & 47&0&+5&+4 &+2& -3\\
\midrule
Concepts  & 64&-1 &-4&-3&  -4     &-& 62&0 &+3&+2&0&-\\
Frames  &  72&0 &-5 &0 &   -1  &-&  63&0 &+1 &+1 &-1&-\\
IgnoreVars  & 79& -1&-1 &-1&  -2   &-&29& +5&+6 &+5&+4&-\\
Named Ent.  &  81&+2& -3&+2&   +3    &-& 50&-18& -9&-7&-9&-\\
Negations  & 87&-1 & -1& +1&  0      &-& 32&-16 & +2& -1&-4&-\\
No WSD  & 78&-1 &0&-1&    -2     &-& 46&+2 &+6&+5&+3&-\\
NS frames &77 & 0&-7&0 &   0    &-&67 & +1&+1&0 &-1&-\\
Reentrancies  & 80&0 &-9&+1&   +2     &-& 48&0 &+2&0&+1&-\\
SRL  & 75&-1&-4&0&  +1  &-& 45&-4&+3&0&+2&-\\
Unlabeled  &75&-1 &0&-1 &   -1  &-&51&0 &+1&+4 &+2&-\\
Wikification  &86&  0 &-2&+1 &  +2  &  -&23& +5  &+6&+6&+7 &-\\
\midrule
$\sum_i\Delta_i$ &\textbf{0} & -5 & \underline{-37} & -5 & -5 & -2 &0& \underline{-25}&\textbf{+27} & +19&+6 &-3\\
%\midrule
%%Smatch  & 47&0&+5&+4 & -3 \\
%Concepts  & 62&0 &+3&+2&-\\
%Frames  &  63&0 &+1 &+1 &-\\
%IgnoreVars  & 29& +5&+6 &+5&-\\
%Named Ent.  & 50&-18& -9&-7&-\\
%Negations  & 32&-16 & +2& -1&-\\
%No WSD  & 46&+2 &+6&+5&-\\
%NS frames &67 & +1&+1&0 &-\\
%Reentrancies  & 48&0 &+2&0&-\\
%SRL  & 45&-4&+3&0&-\\
%Unlabeled  &51&0 &+1&+4 &-\\
%Wikification  &23& +5  &+6&+6 &-\\
\bottomrule

%s & 0.78 & 0.77 & 0.75 & 0.77&0.79\\
\end{tabular}}
\caption{$\rho$ correlation (F1) differences  %with respect to
over different setups (columns), 
%different 
test sets (out-of-domain, in-domain) and %different
subtasks (rows). $\pm x$: plus and minus $x$ pp.$\rho$. }
\label{tab:ablations}
\end{table}

%Hierarchical vs non hierachical, 2 LSTm layers vs 1 LSTm layer. Only words as input (no dependency). No pointers. No Senses. 

%\subsection{AMR and Dependency Relation Investigation}

\section{Conclusion}

AMR parser evaluation with human gold annotation is very costly. %In fact, it is one of the most time-intensive annotation tasks in NLP. 
Our main contributions in this work are two-fold: Firstly, we introduced the concept of automatic AMR accuracy prediction.
%for
Given only an automatic parse and the sentence, from whence it was derived, the goal is to predict evaluation metrics cheaply and possibly at runtime. 
Secondly, we framed the task as a multiple-output regression task and developed a hierarchical neural model %optimized under the multi-task paradigm 
to predict a rich suite of 
%the various kinds of 
AMR evaluation metrics. We 
presented
%have conducted 
three 
%different 
case studies proving (i) the feasibility of automatic AMR accuracy prediction in general (significant correlation with gold scores on unseen in-domain and out-of-domain
%ing 
data) and (ii) the applicability %and usefulness of 
of our model
%such method for 
in two 
%different 
use cases. In the first study, we ranked different automatic candidate parses per sentence, outperforming the random selection baseline by 5.7 pp.\ average Smatch F1 (in-domain) and 5.2 pp.\ (out-of-domain). In the second study, we ranked team submissions to two AMR shared tasks and our method was able to reproduce rankings similar to the true rankings.

\section*{Acknowledgments}
%This work has been supported by the German Research Foundation as part of the Research Training Group Adaptive Preparation of Information from Heterogeneous  Sources  (AIPHES)  under  grant No.\ GRK 1994/1 and by  the  Leibniz  ScienceCampus ``Empirical  Linguistics  and  Computational  Language  Modeling'', supported by the Leibniz Association grant no.  SAS-2015-IDS-LWC  and  by  the  Ministry  of  Science, Research, and Art of Baden-W\"urttemberg. We thank the NVIDIA Corporation for donating the GPUs used in this research.
This work has been supported by the German Research Foundation (grant no.\ GRK 1994/1) and the Leibniz Association (grant no.\ SAS-2015-IDS-LWC)  and the  Ministry  of  Science, Research, and Art of Baden-W\"urttemberg. We are grateful to the NVIDIA corporation for donating the GPU used in this research.

\bibliography{naaclhlt2019}
\bibliographystyle{acl_natbib}
\appendix
\section{Supplemental Material}
\label{sec:supplemental}

\paragraph{Hyper parameters and weights initialization} We initialize all parameters of the model randomly. Embedding vectors of dimension 128 are drawn from $U(0.05,0.05)$ and the LSTM  weights (neurons: 128) and weights of the feed forward output layers are sampled from a Glorot uniform distribution \cite{DBLP:journals/jmlr/GlorotB10}.  For future work, initializing the embedding layer with pre-trained vectors could further increase the performance. In this work, however, we learn all parameters from the given data. %In this work, we optimize all parameters based on the given data alone. %Renouncing pre-trained embeddings, hoewever, has the advantage that all the model's parameters are learned from the given data alone. 
We fit our model using Adam \cite{DBLP:journals/corr/KingmaB14} (learning rate: 0.001) on the training data over 20 epochs with mini batches of size 16. We apply early stopping according to the maximum Pearson's $\rho$ (with regard to Smatch F1) on the development data.
$    \rho =\frac{\sum ^n _{i=1}(x_i - \bar{x})(y_i - \bar{y})}{\sqrt{\sum ^n _{i=1}(x_i - \bar{x})^2} \sqrt{\sum ^n _{i=1}(y_i - \bar{y})^2}}$
quantifies the linear relationship between predicted scores ($x_1,...,x_n$) and true scores ($y_1,...,y_n$).
%Submissions may include non-readable supplementary material used in the work and described in the paper. Any accompanying software and/or data should include licenses and documentation of research review as appropriate. Supplementary material may report preprocessing decisions, model parameters, and other details necessary for the replication of the experiments reported in the paper. Seemingly small preprocessing decisions can sometimes make a large difference in performance, so it is crucial to record such decisions to precisely characterize state-of-the-art methods. 

%Nonetheless, supplementary material should be supplementary (rather
%than central) to the paper. {\bf Submissions that misuse the supplementary 
%material may be rejected without review.}
%Supplementary material may include explanations or details
%of proofs or derivations that do not fit into the paper, lists of
%features or feature templates, sample inputs and outputs for a system,
%pseudo-code or source code, and data. (Source code and data should
%be separate uploads, rather than part of the paper).

%The paper should not rely on the supplementary material: while the paper
%may refer to and cite the supplementary material and the supplementary material will be available to the
%reviewers, they will not be asked to review the
%supplementary material.

\end{document}